\documentclass[letterpaper, 10 pt, conference]{ieeeconf}  % Comment this line out if you need a4paper

\IEEEoverridecommandlockouts                              % This command is only needed if 
\usepackage{graphicx} % for including graphics
\usepackage{amsmath} % assumes amsmath package installed
\usepackage{amssymb}  % assumes amsmath package installed
\usepackage{booktabs}
\usepackage{bm}

\title{\LARGE \bf
HiFAR: Multi-Stage Curriculum Learning for \underline{Hi}gh-Dynamics Humanoid \underline{Fa}ll \underline{R}ecovery
}
\author{Penghui Chen$^{1}$, Yushi Wang$^{1}$, Changsheng Luo$^{1}$, Wenhan Cai$^{2}$, and Mingguo Zhao$^{1}$% <-this % stops a space
\thanks{*This research was supported by STI 2030—Major Projects grant number 2021ZD0201402 and Beijing Natural Science Foundation(L243004).}% <-this % stops a space
\thanks{$^{1}$Department of Automation, Tsinghua University, Beijing 100084, China
 }%
\thanks{$^{2}$Booster Robotics Technology Co., Ltd, Beijing, China}
}

\begin{document}

\maketitle
\thispagestyle{empty}
\pagestyle{empty}

%%%%%%%%%%%%%%%%%%%%%%%%%%%%%%%%%%%%%%%%%%%%%%%%%%%%%%%%%%%%%%%%%%%%%%%%%%%%%%%%
\begin{abstract}

Humanoid robots encounter considerable difficulties in autonomously recovering from falls, especially within dynamic and unstructured environments. Conventional control methodologies are often inadequate in addressing the complexities associated with high-dimensional dynamics and the contact-rich nature of fall recovery. Meanwhile, reinforcement learning techniques are hindered by issues related to sparse rewards, intricate collision scenarios, and discrepancies between simulation and real-world applications. In this study, we introduce a multi-stage curriculum learning framework, termed HiFAR. This framework employs a staged learning approach that progressively incorporates increasingly complex and high-dimensional recovery tasks, thereby facilitating the robot's acquisition of efficient and stable fall recovery strategies. Furthermore, it enables the robot to adapt its policy to effectively manage real-world fall incidents. We assess the efficacy of the proposed method using a real humanoid robot, showcasing its capability to autonomously recover from a diverse range of falls with high success rates, rapid recovery times, robustness, and generalization. 

\end{abstract}

%%%%%%%%%%%%%%%%%%%%%%%%%%%%%%%%%%%%%%%%%%%%%%%%%%%%%%%%%%%%%%%%%%%%%%%%%%%%%%%%
\section{Introduction} 

As robots become increasingly integrated into human-centric spaces, ensuring their resilience and adaptability is essential. Given that falls are inevitable in dynamic environments, the autonomous fall recovery capability of humanoid robots is crucial for real-world deployment.

The need for rapid fall recovery in humanoid robots is driven by the dynamic and unpredictable nature of real-world environments. Quick recovery from falls minimizes downtime and potential damage, enhancing the robot's operational efficiency and safety. For instance, in competitive environments like RoboCup \cite{robocup}, quick and efficient fall recovery is crucial to maintain performance. 

Early solutions for fall recovery relied on finite state machines (FSMs) to trigger preprogrammed actions upon detecting fall indicators \cite{kanehiro2003first}, \cite{stuckler2006getting}. While computationally lightweight—enabling real-time execution on early humanoids like Honda's ASIMO \cite{sakagami2002intelligent}, their brittleness became evident in unstructured environments. 

The need for real-time adaptation in dynamic environments spurred the adoption of optimization-based methods, particularly for maintaining bipedal stability under external perturbations. A seminal advancement in this domain is the Capture Point theory \cite{pratt2006capture}, which formulates push recovery as a problem of regulating the robot’s zero-moment point (ZMP) within its support polygon. By solving convex optimization problems, methods like Model Predictive Control (MPC) \cite{katayama2023model} enabled humanoids to recover balance through step adjustment and torso reorientation. In the context of robot fall recovery, optimization-based methods, such as Whole-Body Control (WBC) \cite{moro2019whole}, have proven effective in online optimization of the robot’s fall trajectory \cite{cai2023self}. However, unmodeled actuator dynamics lead to trajectory tracking errors, and computational latency becomes intractable for multi-contact scenarios.

Model-based control approaches, while effective for stable locomotion, struggle to address the high-dimensional dynamics and contact-rich nature of fall recovery due to their reliance on precise system modeling and computational latency. Reinforcement learning (RL) has been proven effective in humanoid locomotion \cite{kumar2022adapting} and emerges as a solution for fall recovery by enabling robots to learn recovery policies through environment interaction rather than explicit physics modeling. Recent advances in RL have shown success in developing long-horizon behaviors \cite{haarnoja2024learning}, whole-body control \cite{cheng2024expressive}, and contact-rich tasks \cite{zhang2024whole}, \cite{dao2024sim}, making it a promising approach for fall recovery.

Various approaches have been proposed to solve fall recovery with reinforcement learning, including predefined trajectories \cite{haarnoja2024learning}, \cite{zhuang2025embrace}, symmetric behaviors \cite{gaspard2024frasa}, key state initialization(KSI) \cite{yang2023learning}, and task stage division \cite{huang2025learning}. However, these methods have limitations in adaptability to different fall scenarios or robustness to external disturbances.

Autonomous fall recovery in dynamic environments is a critical challenge for humanoid robots. 
This work aims to develop a robust policy that enables a fallen humanoid robot to autonomously recover and stand up across various fall scenarios, including supine, prone, and lateral falls. 
To systematically evaluate the proposed fall recovery policy, we employ the following criteria:
\begin{itemize}
        \item \textbf{Success Rate}: The percentage of successful recovery trials.
        \item \textbf{Recovery Time}: The time required to recover from a fall.
        \item \textbf{Robustness}: The ability to recover from falls under external disturbances and uncertainties.
        \item \textbf{Generalization}: The ability to recover from falls under different initial states and environmental conditions.
\end{itemize}

To address these challenges, we introduce a multi-stage curriculum learning framework that progressively refines the fall recovery policy by increasing task dimensionality and complexity. In the initial stage, the policy is trained in a low-dimensional setting to establish fundamental recovery behaviors. Subsequently, the second stage extends to a high-dimensional deployment scenario, incorporating additional constraints and variability. Each stage integrates tailored optimization techniques to enhance policy robustness and stability.

Our key contributions include:  
\begin{itemize}
        \item We introduce a stage division strategy and curriculum setting to break down the complex high-dimensional fall recovery task into simpler low-dimensional tasks and gradually increase the task complexity to facilitate the learning process.
        \item We employ KSI and reward shaping to guide the learning process and accelerate the convergence of a stable fall recovery policy. Furthermore, we introduce dimensionality expansion through supplementary actuated joints, enabling the policy to generalize robustly across diverse fall scenarios.
        \item We conduct comprehensive experimental validation on the real humanoid robot Booster T1, which illustrates the versatility and robustness of the proposed approach. Real robot experiment videos are available on our project page\footnote{https://hi-far.github.io/}.
\end{itemize}

\section{Related Works}

\subsection{Learning Real-world Humanoid Fall Recovery} 
Tuomas Haarnoja \textit{et al.} trained a DRL agent with predefined keyframes on a kidsize humanoid\cite{haarnoja2024learning}. However, forcing the robot to follow certain trajectories limited the fall recovery behaviors. 

FRASA \cite{gaspard2024frasa} integrates fall recovery and stand-up strategies for the Sigmaban humanoid \cite{allali2019rhoban} into a unified framework but simplifies the problem to a planar scenario and enforces symmetric behaviors, which limits the agent's capabilities. 

By using a KSI method, Chuanyu Yang \textit{et al.} proposed a DRL framework that can learn versatile fall recovery policies for different humanoid and quadruped robots in simulated environments. However, this framework has only been validated on the real Jueying Pro quadruped robot \cite{yang2023learning}. 

Ziwen Zhuang \textit{et al.} achieved humanoid standing through mobile manipulation guided by high-level motion commands, yet their approach lacks autonomous recovery capabilities. \cite{zhuang2025embrace}. 

Furthermore, Tao Huang \textit{et al.} proposed the HoST \cite{huang2025learning} method to train fall recovery by segmenting long-horizon motions into multiple short-horizon sub-tasks. However, this explicit segmentation may limit the agent's adaptability to diverse fall scenarios, such as prone and lateral falls.

HumanUp \cite{he2025learning} also utilizes a two-stage curriculum learning framework to train humanoid robot stand-up. However, their second stage is a tracking task of a slowed-down trajectory discovered in the first stage, which limits the agent's recovery speed and adaptability to different fall scenarios. 

\subsection{Multi-Stage Curriculum Learning}
Multi-Stage Curriculum Learning (MSCL) represents a training methodology that incrementally escalates the complexity of learning tasks by introducing supplementary constraints or objectives at each phase \cite{bengio2009curriculum}. 
Empirical evidence suggests that multi-stage learning and curriculum learning enhances the sample efficiency and generalization capabilities of reinforcement learning agent \cite{wang2024multi}, \cite{chen2024u}, particularly within complex and high-dimensional environments such as fall recovery training \cite{tao2022learning}. 

\subsection{State Initialization}
Reference state initialization (RSI) \cite{peng2018deepmimic} samples the initial state of the agent from the reference motion at the start of each episode. KSI \cite{yang2023learning} initializes the agent's state to a predefined key state, while do not require a reference motion sequence.

RSI leverages the rich and informative state distribution of reference motion to guide the agent during training. KSI contributes to the stabilization of the learning process and enhances the convergence of the policy by offering the agent multiple consistent starting points.

\subsection{Reinforcement Learning Framework}

Many open-source reinforcement learning frameworks are available for humanoid robots, such as Legged Gym and Humanoid Gym \cite{gu2024humanoid}. 
We leverage Booster Gym \cite{BoosterGym} as the foundational codebase to implement our methods. It incorporates pre-configured settings and techniques for sim-to-real transfer, along with a comprehensive toolchain supporting policy evaluation across multiple simulators and real-world deployment. This framework collectively reduces engineering overhead while accelerating the development of our methods.

\begin{figure*}[ht]
        \centering
        \includegraphics[width=\textwidth]{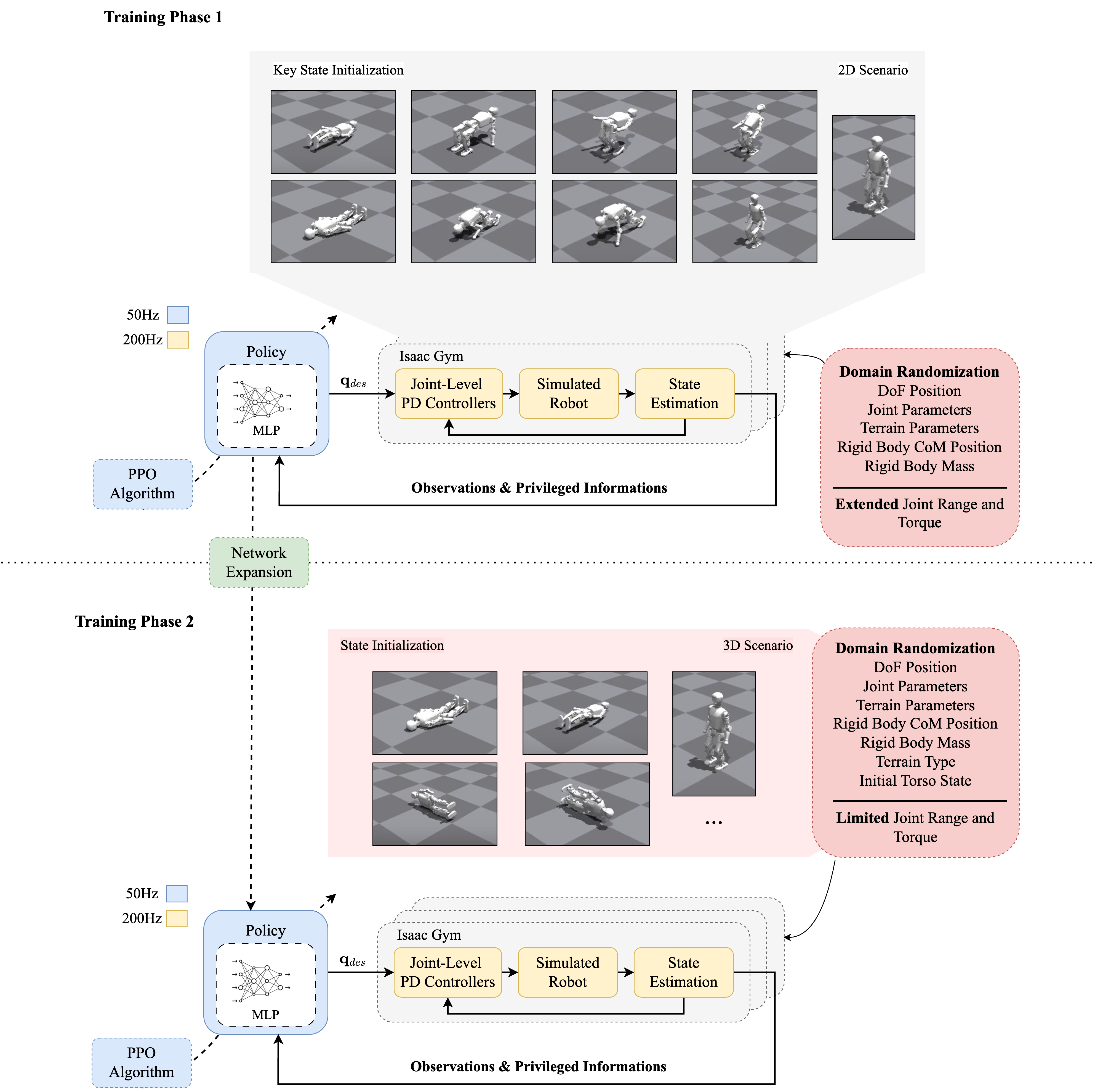}
        \caption{\textbf{Framework overview}. HiFAR offers a multi-stage curriculum learning framework designed for autonomous fall recovery in humanoid robots. The initial training phase emphasizes the development of a fall recovery policy within a low-dimensional task environment. The subsequent training phase tackles the intricacies associated with formulating a deployable fall recovery policy in a higher-dimensional task setting. 
 }
        \label{fig:roadmap_figure}
\end{figure*}

\section{Methods}

\subsection{Framework and Curriculum Setup}

The high-dimensional nature of the fall recovery task, coupled with the complexities of collisions and contacts, as well as the sparse reward structure, presents significant challenges in \textbf{the development of an effective fall recovery policy}. The variability of fall scenarios encountered in real-world applications, along with hardware constraints and the discrepancies between simulated environments and actual conditions, further complicate the learning and implementation of such policies, thereby hindering \textbf{the training of a deployable fall recovery strategy}. Consequently, the direct application of single-stage reinforcement learning to develop an autonomous and robust fall recovery policy proves to be problematic.

To mitigate these challenges, we propose a multi-stage curriculum learning framework, referred to as HiFAR, as illustrated in Fig. \ref{fig:roadmap_figure}. This framework incrementally introduces the agent to increasingly complex and high-dimensional recovery tasks. The multi-stage curriculum learning framework is divided into two distinct stages:

The initial stage concentrates on \textbf{the development of a basic fall recovery policy}. In this phase, the task is constrained to two-dimensional fall scenarios (\textit{e.g.}, supine and prone falls), with control limited to joints operating within the $(x, z)$ plane (the joints highlighted in yellow in Fig. \ref{fig:target_state}(A)). This design choice is based on the observation that humanoid robots can effectively recover from these standard positions using predominantly planar movements. By constraining the policy's exploration space, this configuration reduces the risk of self-collision during training. The agent is trained to recover from simple falls by utilizing an extended range of motion and higher torque capacities. Techniques such as KSI and reward shaping are employed to facilitate the learning process.

The subsequent stage focuses on \textbf{the training of a deployable fall recovery policy}. This phase expands the task to encompass three-dimensional fall scenarios, such as cross-leg falls. To handle diverse real-world falls, we include lateral hip roll joints (orange in Fig. \ref{fig:target_state}(A)) in the action space and introduce further randomizations in terrain and robot state. Realistic constraints on joint positions, torque, and velocity ensure practical training.

The curriculum across the two stages includes the following components:
\begin{itemize}
        \item \textbf{Task Dimension}: The initial stage is centered on a two-dimensional task, whereas the subsequent stage progresses to a three-dimensional task.
        \item \textbf{Task Constraints}: The first stage is characterized by lenient constraints that promote exploration, in contrast to the second stage, which imposes stringent constraints and introduces complex perturbations.
        \item \textbf{Task Complexity}: The initial stage is concerned with straightforward fall scenarios, while the second stage incorporates a greater variety of randomized fall scenarios.
\end{itemize}

By first mastering simpler tasks, the agent is better prepared to tackle more complex challenges, ultimately resulting in more efficient and stable strategies for fall recovery.

\subsection{Training Setup}
\begin{figure}[h!]
        \centering
        \includegraphics[width=\columnwidth]{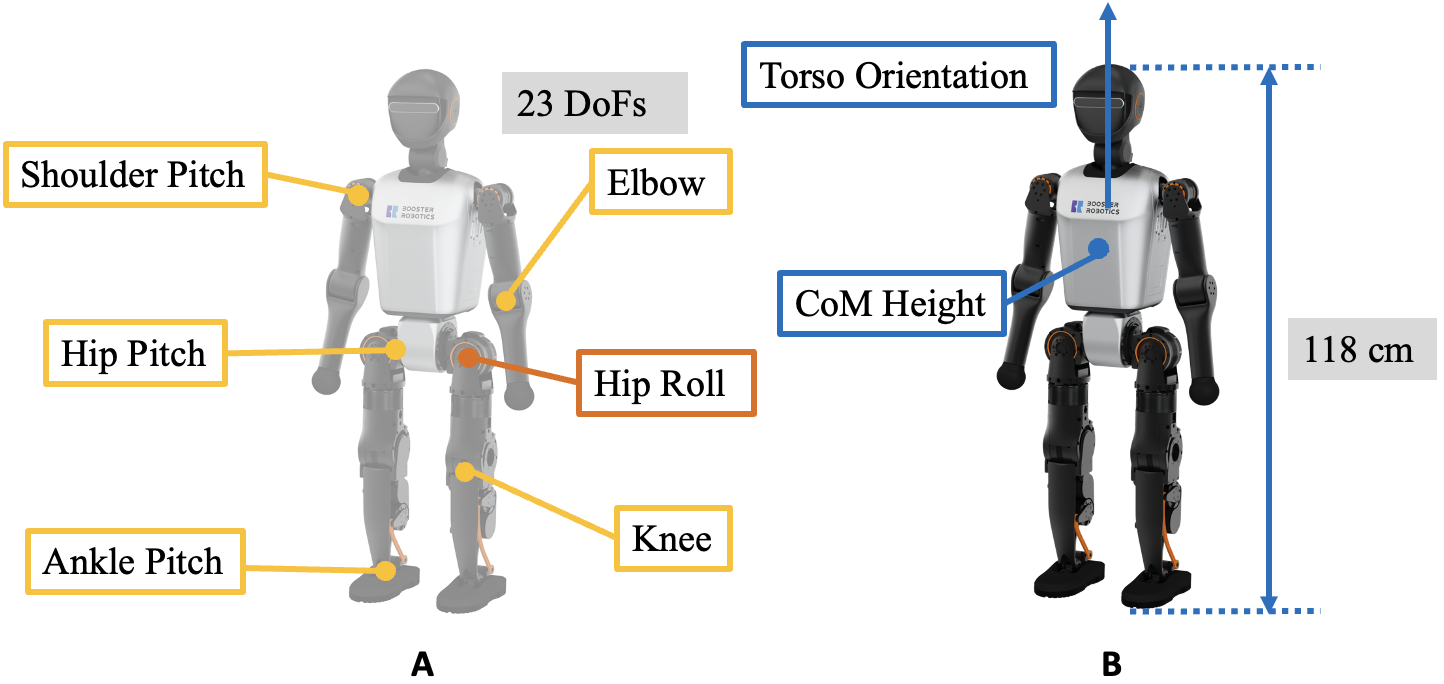}
        \caption{\textbf{Booster T1}. (A) Controlled joints. (B) Target Standing State. }
        \label{fig:target_state}
\end{figure}

The agent is trained in a simulated environment using the Booster Gym framework \cite{BoosterGym}. The target state of fall recovery is the robot standing up with a stable posture, defined by target CoM height, DoF positions, and torso orientation, as shown in Fig. \ref{fig:target_state}(B).

The agent has proprioceptive observations, including the robot's orientation, angular velocity, joint positions, velocities, and action. The action space consists of the target joint position of controlled joints. We use the Proximal Policy Optimization (PPO) algorithm \cite{schulman2017proximal} and RMA network architecture \cite{kumar2022adapting} to train the agent. 

We achieve cross-stage network expansion by integrating zero-initialized fully-connected layers into both input and output layers of the pre-existing network, thereby enabling the second stage to possess expanded action decision spaces and enhanced environmental perception dimensions while preserving knowledge transfer from the initial training phase.

Instead of forcing symmetric actions like FRASA\cite{gaspard2024frasa}, we use a mirror loss term to encourage the agent to learn symmetric recovery strategies while allowing for asymmetry when necessary, which is more realistic and generalizable to real-world scenarios.

\subsection{Keference State Initialization}

As illustrated in Figure \ref{fig:roadmap_figure}, we have selected six critical states—comprising three prone keyframes and three supine keyframes—from established handcrafted fall recovery motions. These motions are specifically designed to facilitate recovery from both supine and prone falls and have demonstrated efficacy in practical applications. The identified key states are used to initialize the agent’s state at the beginning of each episode, ensuring a consistent starting point for the learning process. This approach stabilizes the training process and improves policy convergence.

\subsection{Reward Design}

Table \ref{table:reward_function} outlines the reward function components, designed to encourage prompt and stable fall recovery. Each component is weighted by a coefficient $\omega_i$ to balance competing objectives, with adjustments across learning stages for task alignment.

Survival, base height, standing, orientation, and DoF reference terms facilitate recovery and upright stability. Conversely, torque, DoF velocity/acceleration, root acceleration, action rate, DoF limits, torque fatigue, power, and hip roll terms penalize undesirable actions, promoting smooth, efficient recovery.

\begin{table}[h!]
        \footnotesize
        \renewcommand{\arraystretch}{1.6}
        \centering
        \caption{Components of the Reward Function}
        \label{table:reward_function}
        \begin{tabular}{l l}
        \toprule
        \textbf{Reward terms} & \textbf{Formulas} \\ \midrule
 Survival & \( R_{\text{survival}} = \bm{1} \) \\ \hline
 Base Height & \( R_{\text{base height}} = \exp\left( -\sigma(h_{\text{base}} - h_{\text{des}})^2 \right) \) \\ \hline
 Standing & \( R_{\text{stand}} = \bm{1}_\text{standing} \) \\ \hline
 Orientation & \( R_{\text{orientation}} = \| \bm{g}_{x,y} \|^2 \) \\ \hline
 DoF Reference & \( R_{\text{dof pos ref}} = \| \bm{q}_{\text{ref}} - \bm{q} \|^2 \) \\ \hline
Torques & \( R_{\text{torques}} = \| \bm{\tau} \|^2 \) \\ \hline
 DoF Velocities & \( R_{\text{dof vel}} = \| \dot{\bm{q}} \|^2 \) \\ \hline
 DoF Accelerations & \( R_{\text{dof acc}} = \| \ddot{\bm{q}} \|^2 \) \\ \hline
 Root Accelerations & \( R_{\text{root acc}} = \| \bm{a} \|^2 \) \\ \hline
 Action Rate & \( R_{\text{action rate}} =  \|\dot{\bm{a}_t}\|^2+\|\ddot{\bm{a}_t}\|^2 \) \\ \hline
 DoF Position Limits & \( R_{\text{dof pos limits}} = \|\bm{1}_\text{out of bound}\| \) \\ \hline
 Torque Fatigue & \( R_{\text{torque fatigue}} = \| \bm{\tau} \oslash \bm{\tau}_{\text{limit}} \|^2 \) \\ \hline
 Power & \( R_{\text{power}} = \| \bm{\tau} \odot \dot{\bm{q}} \| \) \\ \hline
 Hip Roll & \( R_{\text{hip roll}} = ( q_{\text{hip roll}} - q_{\text{ref}} )^2 \) \\ \bottomrule
        \end{tabular}
\end{table}
        
\subsection{Domain Randomization}
We apply domain randomization techniques, including randomized push forces and torques, within the simulation environment to improve the robustness and generalizability of the learned policy. The randomized parameters are detailed in Table \ref{table:domain_randomization}:

\begin{table}[ht]
        \footnotesize
        \renewcommand{\arraystretch}{1.6}
        \centering
        \caption{Domain Randomization Parameters}
        \label{table:domain_randomization}
        \begin{tabular}{l l l l}
        \toprule
        \textbf{Parameter} & \textbf{Range} & \textbf{Operation} & \textbf{Distribution} \\ \midrule
 DoF position & [0., 0.05] & Additive & Gaussian \\ \hline
 Base XY Position & [-1., 1.] & Additive & Uniform \\ \hline
 Base Linear Velocity & [0., 0.1] & Additive & Gaussian \\ \hline
 Joint Stiffness & [0.95, 1.05] & Scaling & Uniform \\ \hline
 Joint Damping & [0.95, 1.05] & Scaling & Uniform \\ \hline
 Terrain Friction & [0.1, 2.0] & Additive & Uniform \\ \hline
 Terrain Compliance & [0.5, 1.5] & Additive & Uniform \\ \hline
 Terrain Restitution & [0.1, 0.9] & Additive & Uniform \\ \hline
 Torso CoM Position & [-0.1, 0.1] & Additive & Uniform \\ \hline
 Torso Mass & [0.8, 1.2] & Scaling & Uniform \\ \hline
 Other CoM Position & [-0.005, 0.005] & Additive & Uniform \\ \hline
 Other Mass & [0.98, 1.02] & Scaling & Uniform \\ \bottomrule
        \end{tabular}
\end{table}

Additionally, Booster Gym \cite{BoosterGym} facilitates the modeling of control delays within the simulation, thereby increasing the realism of the training process and enhancing the applicability of the learned policy to real-world scenarios.

\section{Results and Analysis}

\subsection{Training Analysis}
\begin{figure}[h]
        \centering
        \includegraphics[width=\columnwidth]{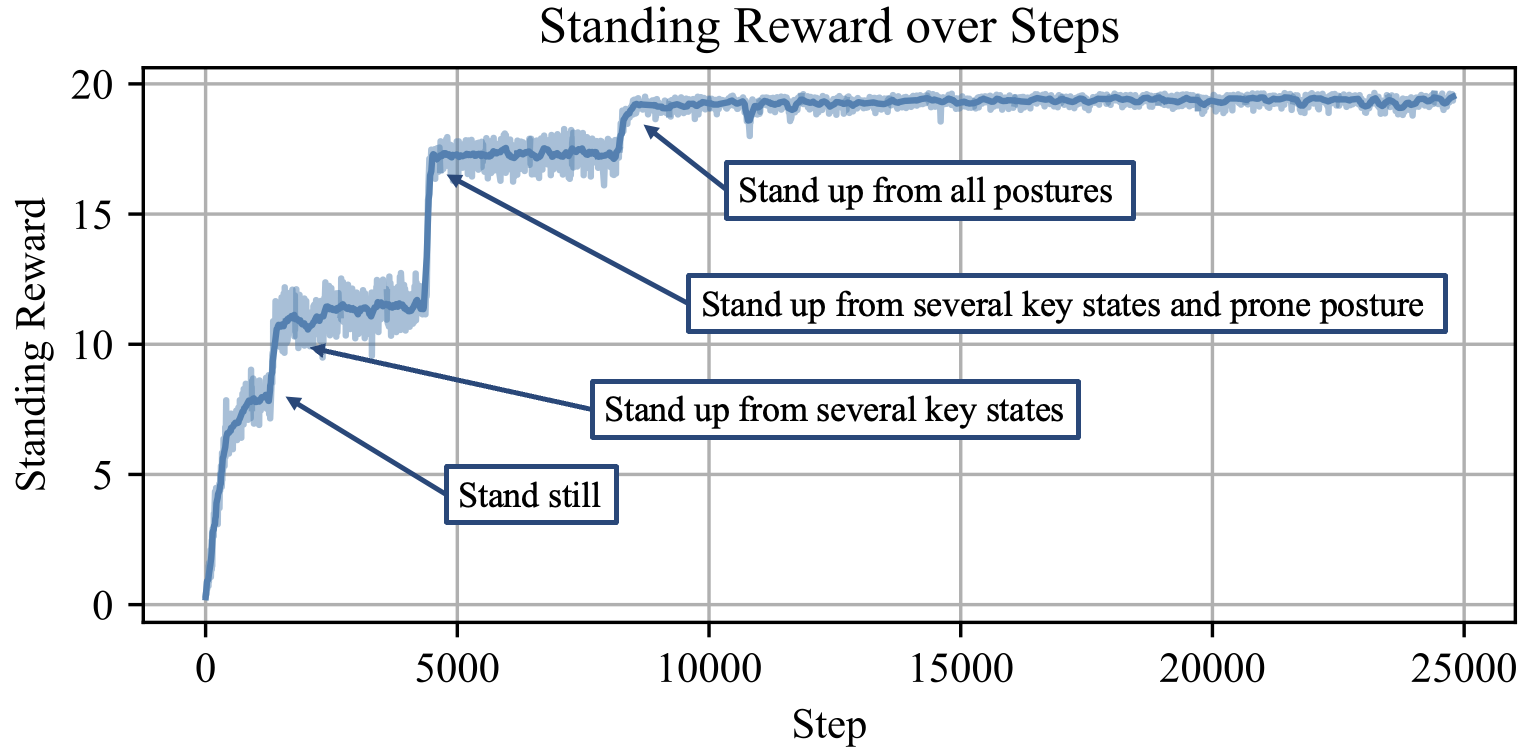}
        \caption{\textbf{Training Analysis}. Standing reward function curve and agent behavior during training.}
        \label{fig:train_stand}
\end{figure}

We analyze the agent’s learning curve with KSI. As shown in Fig. \ref{fig:train_stand}, the standing reward curve exhibits four step-like increases, indicating that the agent progressively learns to recover from falls and stand up at each key state.
The integration of KSI with selected key states stabilizes the learning process and enhances policy convergence.

\subsection{Simulation Experiments}

\subsubsection{Simulation Settings}

We evaluate our policy in the Webots simulator based on success rate, recovery time, and robustness. The simulation environment is designed to replicate real-world conditions, including the robot model and control system.

Before deploying the policy on the physical robot, we conduct a series of simulation experiments to assess its effectiveness in fall recovery. The standard experimental setup includes supine and prone fall scenarios, with a torso mass of 11.7 kg, a friction coefficient of 1.0, and no external disturbances. Each experiment is repeated 10 times under identical conditions to measure the recovery success rate.

\subsubsection{Standard Supine and Prone Recovery Experiments}
\begin{figure*}[h!]
        \centering
        \includegraphics[width=\textwidth]{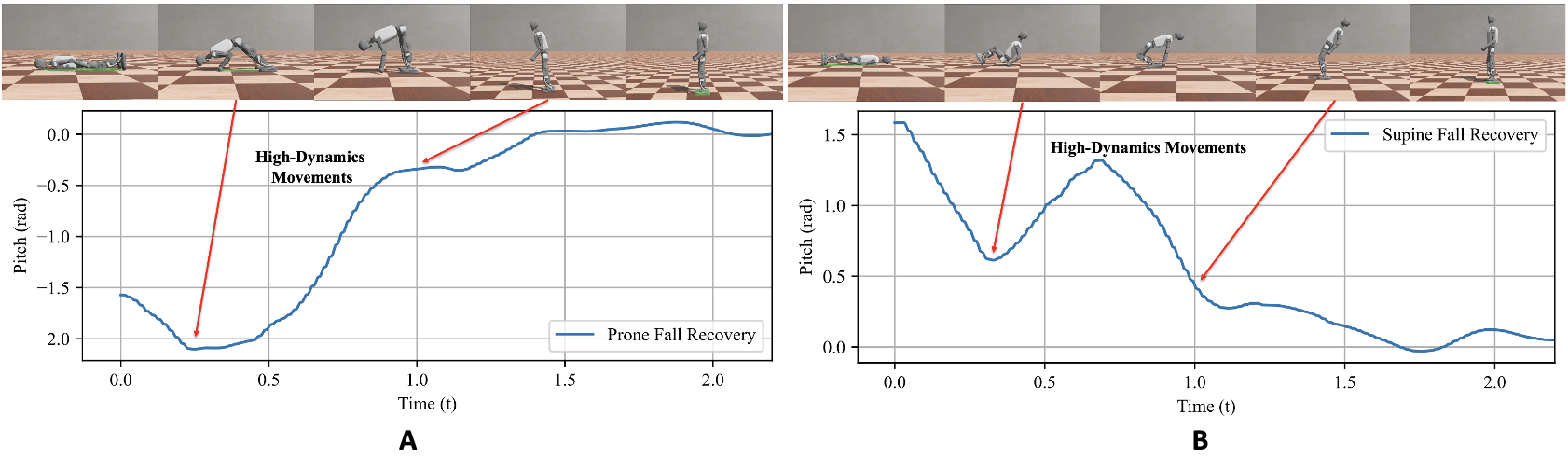}
        \caption{\textbf{Snapshots of simulation experiments and the curve of the torso pitch}. (A) Prone fall recovery. (B) Supine fall recovery.}
        \label{fig:webots}
\end{figure*}

Snapshots of the simulation experiments are presented in Fig. \ref{fig:webots}. In both prone and supine scenarios, our policy successfully enables the robot to recover from falls and stand upright with a stable posture, achieving a 100\% success rate.
The rapid variations in the torso pitch curve highlight the robot's highly dynamic movements.

\subsubsection{Disturbance Experiments}
In this experiment, we apply a random push force to the robot at 0.8s after the recovery process starts and lasts 200ms. 

In the planar force experiment, the perturbation force is applied forward along the robot’s torso direction during recovery from the prone position and backward during recovery from the supine position. The recovery success rate under different perturbation forces is presented in Table \ref{table:force_experiment}.

\begin{table}[h!]
        \footnotesize
        \renewcommand{\arraystretch}{1.6}
        \centering
        \caption{Success Rate of Planar Force Experiments}
        \label{table:force_experiment}
        \begin{tabular}{c c c c c c c}
        \toprule
 \textbf{Force} & \textbf{50N} & \textbf{100N} & \textbf{150N} & \textbf{200N} & \textbf{250N} & \textbf{300N} \\ \midrule
 Prone & $100\%$ & $100\%$ & $100\%$ & $100\%$ & $80\%$ & $0\%$  \\ \hline
 Supine & $100\%$ & $100\%$ & $100\%$ & $100\%$ & $100\%$ & $60\%$ \\ \bottomrule
 \end{tabular}
\end{table}

In the lateral force experiment, the perturbation force is applied leftward along the robot’s torso direction during recovery from both prone and supine positions. The recovery success rate under different perturbation forces is shown in Table \ref{table:lateral_force_experiment}.

\begin{table}[h!]
        \footnotesize
        \renewcommand{\arraystretch}{1.6}
        \centering
        \caption{Success Rate of Lateral Force Experiments}
        \label{table:lateral_force_experiment}
        \begin{tabular}{c c c c c c}
        \toprule
\textbf{Force} & \textbf{50N} & \textbf{100N} & \textbf{150N} & \textbf{200N} & \textbf{250N} \\ \midrule
 Prone & $100\%$ & $100\%$ & $100\%$ & $90\%$ & $10\%$ \\ \hline
 Supine & $100\%$ & $100\%$ & $100\%$ & $100\%$ & $0\%$ \\ \bottomrule
 \end{tabular}
\end{table}

The results demonstrate the strong robustness of our policy against perturbations in various directions.

\subsubsection{Load Experiments}

We simulate the mass inaccuracy of the real robot by modifying its mass in simulation. The success rate of the recovery process under different mass perturbations is shown in Table \ref{table:load_experiment}.

\begin{table}[h!]
        \footnotesize
        \renewcommand{\arraystretch}{1.6}
        \centering
        \caption{Success Rate of Load Experiments}
        \label{table:load_experiment}
        \begin{tabular}{c c c c c}
        \toprule
 \textbf{Extra Torso Mass}  & \textbf{+2.4kg} & \textbf{5.9kg} & \textbf{+9.4kg} & \textbf{+11.7kg} \\ \midrule
 Prone & $100\%$ & $100\%$ & $100\%$ & $100\%$ \\ \hline
 Supine & $100\%$ & $100\%$ & $100\%$ & $0\%$ \\ \bottomrule
 \end{tabular}
\end{table}

With up to $80\%$ extra torso mass, the robot can still recover from prone and supine falls with a $100\%$ success rate, demonstrating the strong robustness of our policy against mass perturbations.

\subsubsection{Friction Experiments}

In real-world scenarios, the robot may fall on terrains with varying friction coefficients. To evaluate its ability to recover from falls under different surface conditions, we conduct experiments with various friction coefficients. The recovery success rate for each friction setting is presented in Table \ref{table:friction_experiment}.

\begin{table}[h!]
        \footnotesize
        \renewcommand{\arraystretch}{1.6}
        \centering
        \caption{Success Rate of Friction Experiments}
        \label{table:friction_experiment}
        \begin{tabular}{c c c c c c}
        \toprule
 \textbf{Friction Coefficient}  & \textbf{0.8} & \textbf{0.6} & \textbf{0.4} & \textbf{0.2} & \textbf{0.1} \\ \midrule
 Prone & $100\%$ & $100\%$ & $100\%$ & $90\%$ & $0\%$ \\ \hline
 Supine & $100\%$ & $100\%$ & $100\%$ & $100\%$ & $90\%$ \\ \bottomrule
 \end{tabular}
\end{table}

The high recovery success rate at friction coefficients as low as 0.2 demonstrates the robustness of our policy and its applicability to real-world terrains with low friction.

\subsubsection{Torque Limits Experiments}

In real-world conditions, motors may be unable to generate the desired torque due to operational constraints. To assess the robot’s ability to recover from falls under varying torque limitations, we conduct experiments with different torque limits. The success rate of the recovery process under different torque limits is shown in Table \ref{table:torque_experiment}.

\begin{table}[h!]
        \footnotesize
        \renewcommand{\arraystretch}{1.6}
        \centering
        \caption{Success Rate of Torque Limit Experiments}
        \label{table:torque_experiment}
        \begin{tabular}{c c c }
        \toprule
 \textbf{Torque Limit Percentage} & $\bm{85\%}$ & $\bm{75\%}$ \\ \midrule
 Prone & $100\%$ & $100\%$ \\ \hline
 Supine & $100\%$ & $60\%$ \\ \bottomrule
 \end{tabular}
\end{table}

The high recovery success rate at an 85\% torque limit demonstrates that the robot can reliably recover from falls despite actuator wear and degradation.

\section{Real-world Experiments and Analysis}
\subsection{Hardware Platform}

We validate our policy on the real humanoid robot Booster T1, which is a 118cm tall humanoid robot with 23 DoFs. The robot is equipped with a set of sensors, including IMUs, and joint encoders to provide feedback for the control system. On-board computations are performed using a Nvidia AGX Orin GPU and 14-core high-performance CPU. 

% \vspace{-0.2cm}

\begin{figure*}[t!]
        \centering
        \includegraphics[width=0.9\textwidth]{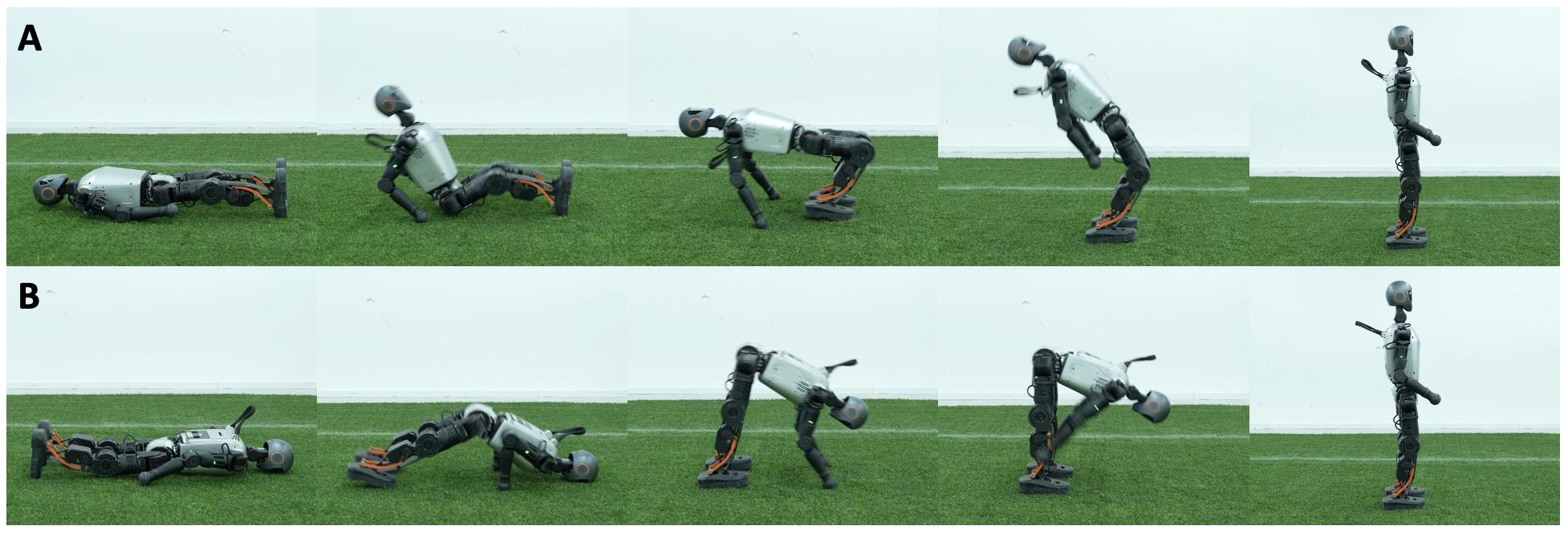}
        \caption{\textbf{Fall recovery behaviors on a real robot}. (A) High-dynamics fall recovery from the prone position. (B) High-dynamics fall recovery from the supine position. }
        \label{fig:result_1}
\end{figure*}

\begin{figure*}[t!]
        \centering
        \includegraphics[width=0.9\textwidth]{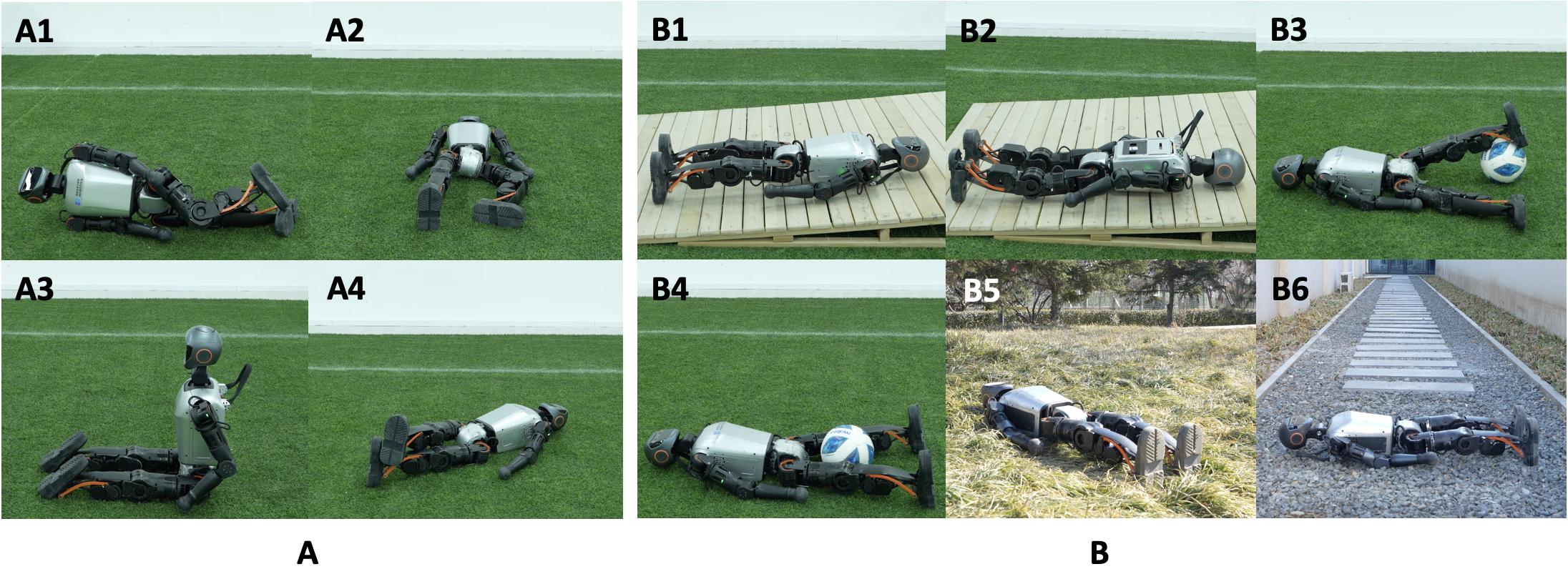}
        \caption{\textbf{Fall recovery under different scenarios}. (A) Random initial state. (A1) lateral positioning, (A2) apart leg posture, (A3) seated position, and (A4) crossed leg posture. (B)Environmental complexity. (B1-2) sloped surfaces, (B3-4) obstacles, and (B5-6) outdoor environments.}
        \label{fig:result_23}
\end{figure*}

\subsection{Experimental Setup}
\subsubsection{Supine and Prone Recovery Experiments} 
Initially placed in either a supine or prone position, the robot executes the policy to restore itself to an upright posture. We record the success rate and recovery time for each trial.

\subsubsection{Random Initial State Experiments} 
Initialized in various fall conditions, including lateral falls, crossed-leg falls, and sitting positions, the robot's ability to recover from diverse initial states is assessed.

\subsubsection{Environmental Complexity Experiments}
Recovery performance is evaluated in complex scenarios, such as standing up from a slope, recovering with obstacles (e.g., a ball) between its legs, and operating in outdoor environments.

\subsubsection{Load and Disturbance Experiments} 
The robot’s recovery ability is tested under additional loads and subjected to strong external forces, such as pushes or impacts, to evaluate its robustness and stability.

\subsection{Results and Analysis}

\begin{figure*}[t!]
        \centering
        \includegraphics[width=0.9\textwidth]{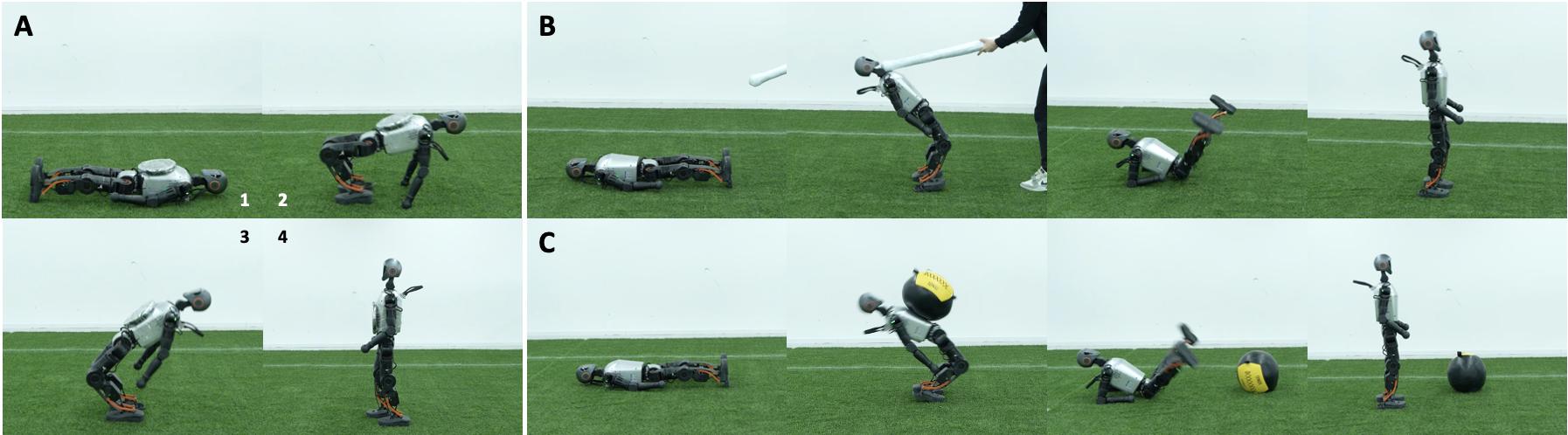}
        \caption{\textbf{Fall recovery under load and disturbance}. (A) 5kg load. (B) Push. (C) Impact.}
        \label{fig:result_4}
\end{figure*}

\subsubsection{Supine and Prone Recovery Experiments}

The performance of our policy is shown in Figure \ref{fig:result_1}. Our policy demonstrates high-dynamics recovery behaviors on the real robot, closely matching its performance in the simulation environment. This consistency validates the sim-to-real transferability of our method.

To evaluate recovery performance, we conducted 20 supine and 20 prone fall recovery trials on the real robot. The average recovery time and success rates are summarized in Table \ref{table:recovery_time_success_rate}. Our method achieves rapid and reliable fall recovery with a 100\% success rate, demonstrating its robustness in real-world deployment.

\begin{table}[h!]
        \footnotesize
        \renewcommand{\arraystretch}{1.6}
        \centering
        \caption{Average Time and Success Rate of Supine and Prone Recovery Experiments}
        \label{table:recovery_time_success_rate}
        \begin{tabular}{c c c}
        \toprule
         & \textbf{Supine} & \textbf{Prone} \\ \midrule
 Average Time (s) & $2.711\pm0.319$ & $2.875\pm0.474$ \\ \hline
 Success Rate & $100\%$ & $100\%$ \\ \bottomrule
        \end{tabular}
\end{table}

\subsubsection{Random Initial State and Environmental Complexity Experiments}

To assess generalization, we initialize the robot in various random states, as illustrated in Figure \ref{fig:result_23}(A). The robot successfully recovers from all scenarios, demonstrating the adaptability of our policy.

Further, we test the robot in complex environments, including standing up from a slope, navigating obstacles between its legs, and recovering in outdoor conditions (Figure \ref{fig:result_23}(B)). The robot consistently executes versatile recovery behaviors across diverse environmental conditions, showcasing the generalizability of our approach.

\subsubsection{Load and Disturbance Experiments}

We evaluate recovery performance under additional physical constraints. Figure \ref{fig:result_4}(A) illustrates the robot with a 5 kg load attached during recovery. The robot successfully stands up while maintaining stability, demonstrating its ability to handle additional payloads.

Furthermore, we introduce external disturbances by applying push forces (Figure \ref{fig:result_4}(B)) and impact (Figure \ref{fig:result_4}(C)) during recovery. The robot reacts by adopting a safe sitting posture to mitigate impact before immediately executing high-dynamic recovery to a standing stance. These results confirm the robustness of our policy in handling strong disturbances autonomously.

\subsubsection{Various Behaviors}

\textbf{Push and Fall Recovery}: The robot uses ankle and hip strategies for minor perturbations and recovery steps for stronger pushes. If falling, it transitions to a safe posture before recovering.
\textbf{Walking after Recovery}: After standing up, the robot stabilizes in a standard stance, enabling a seamless transition into locomotion, demonstrating the practicality of our approach.

\section{Conclusions}

We introduce a multi-stage curriculum learning approach for autonomous humanoid fall recovery. Our method progressively increases task complexity, starting with basic recovery skills and advancing to complex scenarios while addressing sim-to-real challenges.

Simulation and real-world experiments confirm our method’s high success rate, fast recovery, robustness, and generalization across diverse falls and disturbances. The approach enables versatile recovery behaviors in varied environments.

This method offers a robust solution for humanoid fall recovery and can extend to other adaptive robotics applications. 
Future work will aim to enhance the policy's adaptability to a wider range of scenarios, particularly in environments where space or speed constraints affect the robot's stand-up ability.

\section*{Acknowledgment}
We thank Booster Robotics for T1 hardware and support.

\bibliographystyle{IEEEtran}
\bibliography{IEEEabrv,reference}

\begin{thebibliography}{10}
\providecommand{\url}[1]{#1}
\csname url@rmstyle\endcsname
\providecommand{\newblock}{\relax}
\providecommand{\bibinfo}[2]{#2}
\providecommand\BIBentrySTDinterwordspacing{\spaceskip=0pt\relax}
\providecommand\BIBentryALTinterwordstretchfactor{4}
\providecommand\BIBentryALTinterwordspacing{\spaceskip=\fontdimen2\font plus
\BIBentryALTinterwordstretchfactor\fontdimen3\font minus \fontdimen4\font\relax}
\providecommand\BIBforeignlanguage[2]{{%
\expandafter\ifx\csname l@#1\endcsname\relax
\typeout{** WARNING: IEEEtran.bst: No hyphenation pattern has been}%
\typeout{** loaded for the language `#1'. Using the pattern for}%
\typeout{** the default language instead.}%
\else
\language=\csname l@#1\endcsname
\fi
#2}}

\bibitem{robocup}
R.~Federation, ``Robocup official website,'' \url{https://www.robocup.org/}, 2025.

\bibitem{kanehiro2003first}
F.~Kanehiro, K.~Kaneko, K.~Fujiwara, K.~Harada, S.~Kajita, K.~Yokoi, H.~Hirukawa, K.~Akachi, and T.~Isozumi, ``The first humanoid robot that has the same size as a human and that can lie down and get up,'' in \emph{2003 IEEE International Conference on Robotics and Automation (Cat. No. 03CH37422)}, vol.~2.\hskip 1em plus 0.5em minus 0.4em\relax IEEE, 2003, pp. 1633--1639.

\bibitem{stuckler2006getting}
J.~St{\"u}ckler, J.~Schwenk, and S.~Behnke, ``Getting back on two feet: Reliable standing-up routines for a humanoid robot.'' in \emph{IAS}.\hskip 1em plus 0.5em minus 0.4em\relax Citeseer, 2006, pp. 676--685.

\bibitem{sakagami2002intelligent}
Y.~Sakagami, R.~Watanabe, C.~Aoyama, S.~Matsunaga, N.~Higaki, and K.~Fujimura, ``The intelligent asimo: System overview and integration,'' in \emph{IEEE/RSJ international conference on intelligent robots and systems}, vol.~3.\hskip 1em plus 0.5em minus 0.4em\relax IEEE, 2002, pp. 2478--2483.

\bibitem{pratt2006capture}
J.~Pratt, J.~Carff, S.~Drakunov, and A.~Goswami, ``Capture point: A step toward humanoid push recovery,'' in \emph{2006 6th IEEE-RAS international conference on humanoid robots}.\hskip 1em plus 0.5em minus 0.4em\relax Ieee, 2006, pp. 200--207.

\bibitem{katayama2023model}
S.~Katayama, M.~Murooka, and Y.~Tazaki, ``Model predictive control of legged and humanoid robots: models and algorithms,'' \emph{Advanced Robotics}, vol.~37, no.~5, pp. 298--315, 2023.

\bibitem{moro2019whole}
F.~L. Moro and L.~Sentis, ``Whole-body control of humanoid robots,'' \emph{Humanoid robotics: a reference}, pp. 1161--1183, 2019.

\bibitem{cai2023self}
Z.~Cai, Z.~Yu, X.~Chen, Q.~Huang, and A.~Kheddar, ``Self-protect falling trajectories for humanoids with resilient trunk,'' \emph{Mechatronics}, vol.~95, p. 103061, 2023.

\bibitem{kumar2022adapting}
A.~Kumar, Z.~Li, J.~Zeng, D.~Pathak, K.~Sreenath, and J.~Malik, ``Adapting rapid motor adaptation for bipedal robots,'' in \emph{2022 IEEE/RSJ International Conference on Intelligent Robots and Systems (IROS)}.\hskip 1em plus 0.5em minus 0.4em\relax IEEE, 2022, pp. 1161--1168.

\bibitem{haarnoja2024learning}
T.~Haarnoja, B.~Moran, G.~Lever, S.~H. Huang, D.~Tirumala, J.~Humplik, M.~Wulfmeier, S.~Tunyasuvunakool, N.~Y. Siegel, R.~Hafner, \emph{et~al.}, ``Learning agile soccer skills for a bipedal robot with deep reinforcement learning,'' \emph{Science Robotics}, vol.~9, no.~89, p. eadi8022, 2024.

\bibitem{cheng2024expressive}
X.~Cheng, Y.~Ji, J.~Chen, R.~Yang, G.~Yang, and X.~Wang, ``Expressive whole-body control for humanoid robots,'' \emph{arXiv preprint arXiv:2402.16796}, 2024.

\bibitem{zhang2024whole}
Q.~Zhang, P.~Cui, D.~Yan, J.~Sun, Y.~Duan, G.~Han, W.~Zhao, W.~Zhang, Y.~Guo, A.~Zhang, \emph{et~al.}, ``Whole-body humanoid robot locomotion with human reference,'' in \emph{2024 IEEE/RSJ International Conference on Intelligent Robots and Systems (IROS)}.\hskip 1em plus 0.5em minus 0.4em\relax IEEE, 2024, pp. 11\,225--11\,231.

\bibitem{dao2024sim}
J.~Dao, H.~Duan, and A.~Fern, ``Sim-to-real learning for humanoid box loco-manipulation,'' in \emph{2024 IEEE International Conference on Robotics and Automation (ICRA)}.\hskip 1em plus 0.5em minus 0.4em\relax IEEE, 2024, pp. 16\,930--16\,936.

\bibitem{zhuang2025embrace}
Z.~Zhuang and H.~Zhao, ``Embrace collisions: Humanoid shadowing for deployable contact-agnostics motions,'' \emph{arXiv preprint arXiv:2502.01465}, 2025.

\bibitem{gaspard2024frasa}
C.~Gaspard, M.~Duclusaud, G.~Passault, M.~Daniel, and O.~Ly, ``Frasa: An end-to-end reinforcement learning agent for fall recovery and stand up of humanoid robots,'' \emph{arXiv preprint arXiv:2410.08655}, 2024.

\bibitem{yang2023learning}
C.~Yang, C.~Pu, G.~Xin, J.~Zhang, and Z.~Li, ``Learning complex motor skills for legged robot fall recovery,'' \emph{IEEE Robotics and Automation Letters}, vol.~8, no.~7, pp. 4307--4314, 2023.

\bibitem{huang2025learning}
T.~Huang, J.~Ren, H.~Wang, Z.~Wang, Q.~Ben, M.~Wen, X.~Chen, J.~Li, and J.~Pang, ``Learning humanoid standing-up control across diverse postures,'' \emph{arXiv preprint arXiv:2502.08378}, 2025.

\bibitem{allali2019rhoban}
J.~Allali, L.~Gondry, L.~Hofer, P.~Laborde-Zubieta, O.~Ly, S.~N’Guyen, G.~Passault, A.~Pirrone, and Q.~Rouxel, ``Rhoban football club--team description paper,'' Technical report, Tech. Rep., 2019.

\bibitem{he2025learning}
X.~He, R.~Dong, Z.~Chen, and S.~Gupta, ``Learning getting-up policies for real-world humanoid robots,'' \emph{arXiv preprint arXiv:2502.12152}, 2025.

\bibitem{bengio2009curriculum}
Y.~Bengio, J.~Louradour, R.~Collobert, and J.~Weston, ``Curriculum learning,'' in \emph{Proceedings of the 26th annual international conference on machine learning}, 2009, pp. 41--48.

\bibitem{wang2024multi}
D.~Wang, C.~Liu, F.~Chang, H.~Huan, and K.~Cheng, ``Multi-stage reinforcement learning for non-prehensile manipulation,'' \emph{IEEE Robotics and Automation Letters}, 2024.

\bibitem{chen2024u}
Y.~Chen, C.~Yang, T.~Hu, X.~Chen, M.~Lan, L.~Cai, X.~Zhuang, X.~Lin, X.~Lu, and A.~Zhou, ``Are u a joke master? pun generation via multi-stage curriculum learning towards a humor llm,'' in \emph{Findings of the Association for Computational Linguistics ACL 2024}, 2024, pp. 878--890.

\bibitem{tao2022learning}
T.~Tao, M.~Wilson, R.~Gou, and M.~Van De~Panne, ``Learning to get up,'' in \emph{ACM SIGGRAPH 2022 conference proceedings}, 2022, pp. 1--10.

\bibitem{peng2018deepmimic}
X.~B. Peng, P.~Abbeel, S.~Levine, and M.~Van~de Panne, ``Deepmimic: Example-guided deep reinforcement learning of physics-based character skills,'' \emph{ACM Transactions On Graphics (TOG)}, vol.~37, no.~4, pp. 1--14, 2018.

\bibitem{gu2024humanoid}
X.~Gu, Y.-J. Wang, and J.~Chen, ``Humanoid-gym: Reinforcement learning for humanoid robot with zero-shot sim2real transfer,'' \emph{arXiv preprint arXiv:2404.05695}, 2024.

\bibitem{BoosterGym}
B.~Robotics, ``Booster gym: A user-friendly reinforcement learning framework for humanoid robot,'' \url{https://github.com/BoosterRobotics/booster\_gym}, 2024.

\bibitem{schulman2017proximal}
J.~Schulman, F.~Wolski, P.~Dhariwal, A.~Radford, and O.~Klimov, ``Proximal policy optimization algorithms,'' \emph{arXiv preprint arXiv:1707.06347}, 2017.

\end{thebibliography}

\end{document}